\documentclass{article}

\usepackage{PRIMEarxiv}
\usepackage[
pdfauthor={derajan},
pdftitle={Improved Genetic Algorithm Based on Greedy and Simulated Annealing Ideas for Vascular Robot Ordering Strategy},
pdfstartview=XYZ,
bookmarks=true,
colorlinks=true,
linkcolor=blue,
urlcolor=blue,
citecolor=blue,
pdftex,
bookmarks=true,
linktocpage=true,   
hyperindex=true
]{hyperref}
\usepackage{multirow}
\usepackage{makecell}
\usepackage{amsmath} 
\usepackage{amssymb}
\usepackage{graphicx}
\usepackage{float}
\usepackage{times}
\usepackage{latexsym}
\usepackage{booktabs}
\usepackage{subfig}
\usepackage{caption}
\usepackage[utf8]{inputenc} 
\usepackage[T1]{fontenc}    
\usepackage{hyperref}       
\usepackage{url}            
\usepackage{booktabs}       
\usepackage{amsfonts}       
\usepackage{nicefrac}       
\usepackage{microtype}      
\usepackage{lipsum}
\usepackage{fancyhdr}       
\usepackage{graphicx}       
\graphicspath{{media/}}     
\usepackage{amsmath,amssymb}
\usepackage{booktabs} 
\usepackage{tabularx}
\usepackage{changepage}
\usepackage[linesnumbered,ruled,vlined]{algorithm2e}
\usepackage{textcomp,marvosym}
\title{Improved Genetic Algorithm Based on Greedy and Simulated Annealing Ideas for Vascular Robot Ordering Strategy
}

\author{
  Zixi Wang  \\
  School of Computing and Artificial Intelligence \\
  Southwest Jiaotong University \\
  Chengdu \\
  \texttt{zixi-wang@outlook.com} \\
  \And
  Yubo Huang*, Xin Lai \\
  School of Civil Engineering \\
  Southwest Jiaotong University \\
  Chengdu \\
  \texttt{\{ybforever, xinlai\}@my.swjtu.edu.cn} \\
  \And
  Yukai Zhang  \\
  School of Information Science and Technology\\
  Southwest Jiaotong University \\
  Chengdu \\
  \And
  Yifei Sheng  \\
  School of Clinical Medicine\\
  North China University of Science and Technology\\ 
  Tangshan\\
  \And
  Peng Lu  \\
  School of Mathematics \\
  Southwest Jiaotong University \\
  Chengdu \\
}

\begin{document}
\maketitle

\begin{abstract}
This study presents a comprehensive approach for optimizing the acquisition, utilization, and maintenance of ABLVR vascular robots in healthcare settings. Medical robotics, particularly in vascular treatments, necessitates precise resource allocation and optimization due to the complex nature of robot and operator maintenance. Traditional heuristic methods, though intuitive, often fail to achieve global optimization. To address these challenges, this research introduces a novel strategy, combining mathematical modeling, a hybrid genetic algorithm, and ARIMA time series forecasting. Considering the dynamic healthcare environment, our approach includes a robust resource allocation model for robotic vessels and operators. We incorporate the unique requirements of the adaptive learning process for operators and the maintenance needs of robotic components. The hybrid genetic algorithm, integrating simulated annealing and greedy approaches, efficiently solves the optimization problem. Additionally, ARIMA time series forecasting predicts the demand for vascular robots, further enhancing the adaptability of our strategy. Experimental results demonstrate the superiority of our approach in terms of optimization, transparency, and convergence speed from other state-of-the-art methods.
\end{abstract}


\section{Introduction}
Medical robotics is a rapidly evolving field that leverages advanced algorithms to unlock the full potential of cutting-edge technologies \cite{Ghith2022DesignAO}. One of these technologies is the vascular robot, which can perform precise and minimally invasive procedures within the human vasculature. Among the various types of vascular robots, the ABLVR vascular robot is a novel and promising technology that consists of a robotic vessel and four operators who can navigate the bloodstream autonomously \cite{Pschel2022RobotassistedTI,Bao2022MultilevelOS}. However, this technology poses a unique challenge in terms of resource allocation and optimization, as the operators require a week-long biological learning process before they can be fully operational, and the robotic vessel needs to be periodically removed for maintenance \cite{Jin2021DevelopmentOA}. Vascular robots need to be fully trained in the vessel boat before they can work.

The maintenance of medical robots is a healthcare resource allocation problem. Navaz et al. \cite{Navaz2021TrendsTA} conducted a comprehensive review, highlighting various approaches to optimizing resource allocation in this context. Their work serves as a foundational reference, summarizing existing methodologies and identifying research gaps. Faccincani et al. \cite{Faccincani2021AssessingHA}.  investigated adaptive resource allocation strategies, emphasizing the need for dynamic models that can adapt to changing conditions. Their research underscores the importance of flexibility in resource allocation to meet the evolving demands of healthcare environments. Zouri et al. \cite{Zouri2019DecisionSF} delved into cost-effective resource allocation models, emphasizing the importance of cost optimization in rehabilitation hospitals. Their study provides insights into the trade-offs between cost and treatment efficiency, a critical consideration in healthcare robotics. Guo et al. \cite{Guo2019StudyOR} focused on system control models for vascular robots and used robust controllers to optimize their performance and improve system stability. Their research contributes to understanding the dynamics of acquiring and maintaining robotic assets for vascular treatments. However, their study does not directly relate to optimizing the birth of robotic assets for vascular therapy, and it is difficult for hospitals to go directly through their methodology to design and optimize acquisition strategies for robots for vascular therapy.

In recent years, the rapid development of computer technology has allowed it to be used in a wide range of applications in the healthcare industry. Pashaei et al. \cite{pashaei2023hybrid} used a hybrid binary COOT algorithm with simulated annealing to search for targeted genes. Pashaei et al. \cite{pashaei2022hybrid} proposed a simulated annealing-based mRMR search method for feature selection in high-dimensional biomedical data. Yu et al. \cite{yu2021reinforcement} explored the role of reinforcement learning in the healthcare domain for health resource allocation and scheduling and health management problems. These applications show that the application of computer technology in the healthcare industry promotes the development of the healthcare industry.

This study aims to address this challenge by developing a comprehensive and adaptive long-term strategy for the acquisition, utilization, and maintenance of ABLVR vascular robots and operators \cite{Pschel2022RobotassistedTI}. This study aims to address this challenge by developing a comprehensive, adaptable, and long-term strategy for the acquisition, use, and maintenance of the ABLVR vascular robot and operator. We designed a genetic algorithm based on improved greed and simulated annealing, along with an ARIMA time-series model optimized by the genetic algorithm, for optimizing the ordering and scheduling of the ABLVR vascular robot's capacity boats and operators to ensure efficient treatment while minimizing costs and addressing potential damage to the robot. Specifically, with the known number of vascular robot uses required per week, the greedy algorithm is first embedded into a genetic algorithm to solve for the optimal number of vessel boats and operators to be purchased per week as an initial solution to the genetic algorithm. Then the genetic algorithm optimized based on the simulated annealing idea is built to solve the final result. The ARIMA model, optimized by the genetic algorithm, forecasts the unknown demand for vascular robot uses in a time series, and determines the optimal number of vessel boats and operators to purchase accordingly. The key contributions and highlights of our research include:
\begin{itemize}
\item [$\bullet$] Comprehensive Resource Allocation Model: We have developed a robust resource allocation model that optimizes the procurement of both robotic vessels and operators, considering the dynamic nature of healthcare environments.
\item [$\bullet$] Incorporating Adaptive Learning: Our model accounts for the adaptive learning process required for operators, as well as the maintenance and disposal of robotic components.
\item [$\bullet$]Hybrid Genetic Algorithm: We introduce a hybrid genetic algorithm that incorporates simulated annealing and greedy approaches to efficiently solve the optimization problem.
\item [$\bullet$]Time Series Forecasting: We use an ARIMA time series model to predict the demand for vascular robots, enhancing the adaptability of our procurement strategy.
\end{itemize}

We compare our proposed method with traditional heuristic approaches and machine learning-based methods, highlighting the advantages of our approach in terms of optimization and transparency. This study proposes an innovative approach based on an improved genetic algorithm that can optimize the ordering and scheduling of the ABLVR vascular robots and operators. By applying computational optimization techniques, this study seeks to enhance the efficiency and cost-effectiveness of this groundbreaking medical technology.

\section{Methodology}
\subsection{Assumptions}

This research employs a quantitative approach grounded in computational modeling and optimization techniques to address the multifaceted challenges posed by the acquisition and utilization of ABLVR vascular robots in healthcare settings.
To maintain the reasonableness and accuracy of the model solution in alignment with the specific problem, it is necessary to introduce the following assumptions:
\begin{itemize}
\item [$\bullet$] Cost-Based Part Disposal: The model adopts a cost-centric approach, whereby any component of the robot is discarded if the cost of maintaining it surpasses the combined cost of purchasing a new part and facilitating the learning process for both new and used parts \cite{Wang2023RealTimeAA}. Additionally, parts are considered for disposal if they are no longer viable for reuse.
\item [$\bullet$] Rounding in Calculations: Throughout the calculation process in this paper, rounding is performed at each step. It is assumed that rounding does not introduce significant deviations or impact the overall reliability of the model's results. This assumption is made to ensure the feasibility of computational solutions \cite{Sarkies2021EffectivenessOK}. 
\item [$\bullet$]Predictable Part Reliability: It is assumed that each component of the robot will not experience unexpected failures due to internal problems during use. This assumption simplifies our model, allowing focus on external factors and optimizing the ordering strategy \cite{Koulaouzidis2023RoboticAssistedSF}. 
\end{itemize}

\subsection{Mathematical Modeling}

\subsubsection{Optimal decision making based on the single-objective genetic algorithm}

First, there is need to establish a robot purchase strategy model to make the lowest cost of purchased operators and container boats under the premise of meeting the hospital treatment demand, and the purchase cost is linearly related to the quantity \cite{Mena2014LeadingPS}. In order to better control costs, this paper takes into account that there may be cases where the cost of maintenance to the robot vessel boat or operator exceeds the cost of purchase of that part and learning of new and used parts, i.e

\begin{equation}
W_O\times P_{Om}>P_O+2P_{Ot}
\label{eq1} 
\end{equation}

\vspace{-14pt}

\begin{equation}
W_C\times P_{Cm}>P_C
\label{eq2} 
\end{equation}

Where $W_O$  and $W_C$ indicate the current number of weeks of maintenance for either operator and vessel, respectively. This document chooses to discard this part when the vessel maintenance cost exceeds the purchase price of a new vessel and the operator maintenance cost exceeds the purchase cost of a new operator and the learning cost of a new or old operator, or when the part was last used.

To determine the objective function, the current total number of operators and vessel boats can be expressed as follows,respectively.

\begin{equation}
N_{Ci}=\sum\limits_{j=1}^i{\left( C_{Bi}-C_{Di} \right)}+N_{C_o}
\label{eq3} 
\end{equation}

\vspace{-10pt}

\begin{equation}
N_{Oi}=\sum\limits_{j=1}^i{\left( C_{Oi}-C_{Oi} \right)}+N_{O_0}
\label{eq4} 
\end{equation}

Where $N_{Ci}$  indicates the total number of vessels owned in week i, $N_{Oi}^{u}$ indicates the total number of operators owned in week i, $C_{Bi}$ indicates the number of vessels purchased in week i, and $C_{Di}$ indicates the number of vessels discarded in week i. $N_{C_0}$ indicates the number of vessels owned at the beginning of week 1, and $N_{O_0}$ indicates the number of operators owned at the beginning of week 1.

Hence the total purchase cost can be expressed as:

\begin{equation}
    \begin{aligned}
minP(C_{Bi}, O_{Bi}, N_{Oi}^g, N_{Oi}^t, N_{Ci}^m, N_{Oi}^m)&=\left(\sum C_{Bi}\right)\times P_C+\left(\sum O_{Bi}\right)\times P_O\\&+\left(\sum N_{Oi}^g+N_{Oi}^t\right)\times P_{Ot}+\left(\sum N_{Ci}^m\right)\times P_{Om}\\& +\left(\sum N_{Oi}^m\right)\times P_{Cm}\
\label{eq5} 
   \end{aligned} 
\end{equation}

Where $P_C$ denotes the unit price per vessel boat, and $P_O$ denotes the unit price per operator, the $P_{Ot}$ denotes the price of one operator training, and $P_{Om}$ denotes the price of one operator maintenance, and  $P_{Cm}$ denotes the total cost required to complete hospital treatment operations.

To determine the constraint conditions the relationship between the total number of operators and vessel boats in week 1 and the number in maintenance is shown below:

\begin{equation}
N_{Ci}=N_{Ci}^{m}+N_{Ci}^{u}
\label{eq6} 
\end{equation}

\vspace{-14pt}

\begin{equation}
N_{Ci}^u=R_i 
\label{eq7} 
\end{equation}

\vspace{-14pt}

\begin{equation}
N_{Oi} =N_{Oi}^m+N_{Oi}^u+N_{Oi}^g+N_{Oi}^t
\label{eq8} 
\end{equation}

\vspace{-14pt}

\begin{equation}
N_{Oi}^{u}=4R_{i}
\label{eq9} 
\end{equation}

\vspace{-14pt}

\begin{equation}
N_{Oi}^{t}=O_{Bi}
\label{eq10} 
\end{equation}

\vspace{-14pt}

\begin{equation}
N_{\partial i}^{g}=\left\lceil\frac{O_{Bi}}{G}\right\rceil
\label{eq11} 
\end{equation}

Where G indicates the number of new operators each skilled operator can instruct, and the total number of operators owned in week i is equal to the sum of the number of operators in maintenance, use, training instruction, and training at that point.

Since the vascular robot must be dismantled after one week of work in the vasculature, the robot's operators within it cannot work again until after 7 days of maintenance \cite{da2008overview}. Therefore, the number of operators under maintenance in week i is equal to four times the number of robots in the hospital's operational requirements in week i-1 minus the number of operators discarded in that week, i.e.

\begin{equation}
N_{Oi}^{m}\ge 4R_{i-1}-O_{Di}
\label{eq12} 
\end{equation}

$R_{i-1}$ denotes the number of robots in demand for hospital operations in week i-1.
Since the newly purchased vessel boats cannot start working until after a week of commissioning, there is: the number of vascular robots required by the hospital in week 1 must be less than the total number of vessel boats owned in week  1 minus the number of vessel boats discarded in week 1. This can be expressed mathematically as:

\begin{equation}
\left\{ \begin{array}{l}
	R_i<N_{Ci-1}-C_{Di}\,\,,\,\,i>1\\
	R_i<N_{C0}-C_{Di}\,\,,\,\,i=1\\
\end{array} \right. 
\label{eq13} 
\end{equation}

where $N_{C0}$ indicates the original number of container boats.
Therefore, the robot buying strategy can be modelled as follows:

\begin{equation}
min\,\,P\left(C_{Bi}, O_{Bi}, N_{Oi}^{g}, N_{Oi}^{t}, N_{Ci}^{m}, N_{Oi}^{m} \right) 
\label{eq14} 
\end{equation}

Assuming that 20\% of the vascular robots in the human body are destroyed each week, so that the total number of vessel boats and operators changes each week, the total number of vessel boats and operators by changing the composition of a set of functions can be expressed recursively as:

\begin{equation}
N_{Ci}=N_{Ci-1}+C_{Bi}-C_{Di}-K\times N_{Ci-1}^{u}
\label{eq15} 
\end{equation}

\vspace{-14pt}

\begin{equation}
N_{Oi}=N_{Oi-1}+O_{Bi}-O_{Di}-K\times N_{Oi-1}^{u}
\label{eq16} 
\end{equation}

where K indicates the percentage of vascular robots destroyed in the human body at this time, here K=20\%.Since the robot's operators do not work again until after 7 days of maintenance \cite{bao2018operation}. Therefore, the number of operators under maintenance in week 1 is greater than equal to four times the number of machines in the hospital's operational demand in week 1 minus the number of operators discarded in that week minus the number of macrophages hit, i.e.

\begin{equation}
N_{Oi}^{m}\geqslant R_{i-1}-O_{Di}-K\times N_{Oi-1}^{u}
\label{eq17} 
\end{equation}

Since the newly purchased vessel boats cannot start working until after a week of commissioning, there is: the number of vascular robots required by the hospital in week i must be less than the sum of the total number of vessel boats owned in week i - 1 minus the number of vessel boats discarded in week i and the number of vessel boats destroyed in week i - 1, which can be expressed mathematically as:

\begin{equation}
\left\{ \begin{array}{l}
	R_i<N_{Ci-1}-C_{Di}-K\times N_{Ci-1}^{u}\,\,,\,\,i>1\\
	R_i<N_{C0}-C_{Di}\,\,,\,\,i=1\\
\end{array} \right.
\label{eq18} 
\end{equation}

where $N_{C0}$  indicates the number of original container boats. Now we need to consider the probability of hitting a macrophage resulting in the complete destruction of the vascular robot, while changing the upper limit of how much each skilled operator can instruct a new operator to learn from G to 20.
At this time, 10\% of the vascular robots in the human body are destroyed, and the total number of vessel boats and operators changes each week as the amount of destruction changes, and the total number of vessel boats and operators can be expressed recursively as:

\begin{equation}
N_{Ci}=N_{Ci-1}+C_{Bi}-C_{Di}-K\times N_{Ci-1}^{u}
\label{eq19} 
\end{equation}

\vspace{-14pt}

\begin{equation}
N_{Oi}=N_{Oi-1}+O_{Bi}-O_{Di}-K\times N_{Oi-1}^{u}
\label{eq20} 
\end{equation}

The number of operators in maintenance is equal to four times the number of machines in the hospital's operational requirements in week i-1 minus the number of operators discarded in that week minus the number of macrophages hit, i.e.

\begin{equation}
N_{Oi}^{m}\geqslant4R_{i-1}-O_{Di}-K\times N_{Oi-1}^{u}
\label{eq21} 
\end{equation}

Where K indicates the percentage of vascular robots destroyed in the human body at this time, here K = 10\%, while the upper limit of what each skilled operator can instruct new operators to learn G changes to 20 \cite{zhao2022remote}.

And for vessel boats, the number of vascular robots needed to have week i hospitals must be less than the sum of the total number of vessel boats owned in week i - 1 minus the number of vessel boats discarded in week i and the number of vessel boats destroyed in week i - 1. The inequality can be expressed as:

\begin{equation}
\left\{ \begin{array}{l}
	R_i<N_{Ci-1}-C_{Di}-K\times N_{Ci-1}^{u}\,\,,\,\,i>1\\
	R_i<N_{C0}-C_{Di}\,\,,\,\,i=1\\
\end{array} \right.
\label{eq22} 
\end{equation}

Where $N_{C0}$ denotes the number of original vessel boats and K denotes the percentage of vascular robots destroyed in the human body at this time, here K=10\%. The specific analytical thought process is shown in the following Fig~\ref{flowchart}.

\begin{figure}[!ht] 
\centering 
\includegraphics[width=1\textwidth]{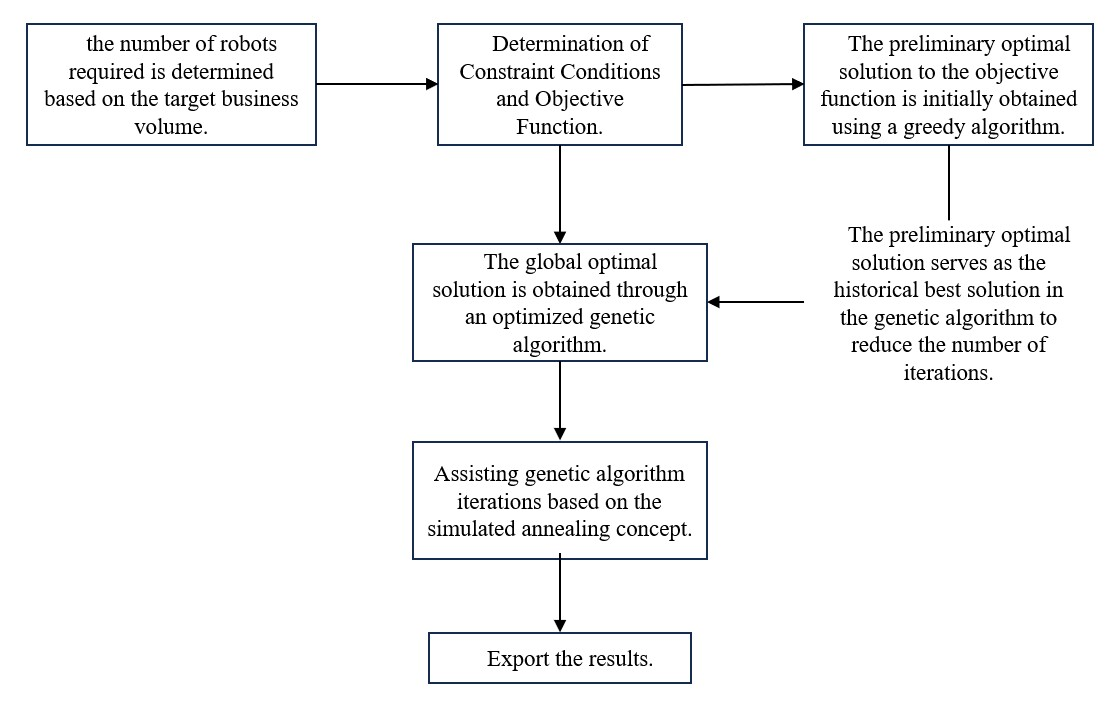} 
\caption{Model Analysis of Optimization Strategies for Vascular Robots.} 
\label{flowchart} 
\end{figure}

\subsubsection{ARIMA Sequences and Seasonal Sequence Forecasting}

In this section, this paper predicts the demand for the use of vascular robots from 105-112 weeks by ARIMA time series \cite{siami2018comparison}. Time series analysis refers to a set of random variables ordered by time. The main idea of the model is to make dynamic predictions of unknown data based on inter-observations by dependence and correlation. This time model is used to forecast the 105-112 week demand by building this time model, and the implementation of the ARIMA model consists of the following five main steps.

\textbf{Step 1}: Perform data processing, for the mean value of the number of vascular robots used in weeks 1-104 in Annex II $y(t)$ can be expressed as \cite{tseng2002combining} :

\begin{equation}
\overline{Y}=\frac{1}{n}\sum\limits_{t=1}^n{y\left( t \right)}
\label{eq23} 
\end{equation}

The sample value of the new sequence obtained after its differential processing can be expressed as:

\begin{equation}
x\left( t \right) =y\left( t \right) -\overline{Y}
\label{eq24} 
\end{equation}

Generally for d-order difference can be expressed as follows: ($\nabla ^d$ is called the d-order difference operator) \cite{xue2016arima}.

\begin{equation}
\nabla ^dX_t=\left( 1-B \right) ^dX_t
\label{eq25} 
\end{equation}

The number of differences is determined by the parameter d in the ARIMA (p,d,0) model, i.e., one difference is made, d=1, i.e., two differences are made, d=2, i.e., no difference is made, at which point the model structure is changed to ARIMA (p).

\textbf{Step 2}: Parameter estimation: Recursive least squares with forgetting factors can be used for parameter estimation.

The forgetting factor enhances the effect of current observations on parameter estimation while weakening the effect of previous observations. The inclusion of the forgetting factor in recursion can take into account the time-varying nature of the model parameters, and the ARIMA (p) model for the sequence $y(t)$ can be expressed as\cite{tseng2002combining}:

\begin{equation}
y\left( t \right) =\boldsymbol{\varphi }^T\left( t \right) \boldsymbol{\theta }+e\left( t \right) 
\label{eq26} 
\end{equation}

\vspace{-14pt}

\begin{equation}
\boldsymbol{\varphi }^T\left( t \right) =\left[ y\left( t-1 \right) ,y\left( t-2 \right) ,\cdots ,y\left( t-p \right) \right] 
\label{eq27} 
\end{equation}

\vspace{-14pt}

\begin{equation}
\boldsymbol{\theta }=\left[ a_1,a_2,\cdots ,a_p \right] ^T
\label{eq28} 
\end{equation}

Recursive parameter estimation by substituting $\boldsymbol{\varphi }^T$, $\boldsymbol{\theta }$ into a recursive least squares formulation with a forgetting factor.

\textbf{Step 3}: Forecasting algorithm: the Astrom forecasting method based on the linear minimum variance forecasting principle which can better solve the random geodesic problem in forecasting is used for forecasting, and the ARIMA (p,d,q) process can be expressed as:

\begin{equation}
A\left( B \right) \nabla ^dy\left( t \right) =C\left( B \right) e\left( t \right) 
\label{eq29} 
\end{equation}

where y(t), e(t) denote the original sequence and the white noise sequence, respectively.

\begin{equation}
A\left( B \right) =1-a_1B-a_2B^2-\cdots -a_pB^p
\label{eq30} 
\end{equation}

\vspace{-14pt}

\begin{equation}
C\left( B \right) =1-c_1B-c_2B^2-\cdots -c_pB^p
\label{eq31} 
\end{equation}

B denotes the back-shift operator is:

\begin{equation}
B^ny\left( t \right) =y\left( t-n \right) ,n=1,2,\cdots 
\label{eq32} 
\end{equation}

Minimum variance predictor is:

\begin{equation}
\widehat{Y}\left( t+kt \right) =\frac{G\left( B \right)}{C\left( B \right)}y\left( t \right) 
\label{eq33} 
\end{equation}

\textbf{Step 4}: model check: this is achieved by checking whether the error series between the original time series and the established model is stochastic; if the model check fails, the model is rebuilt.

\textbf{Step 5}: Export the appropriate prediction model and perform the actual prediction analysis.

\section{Improved Genetic Algorithm Based on Greedy and Simulated Annealing Ideas}

\begin{figure}[!ht] 
\centering 
\includegraphics[width=1\textwidth]{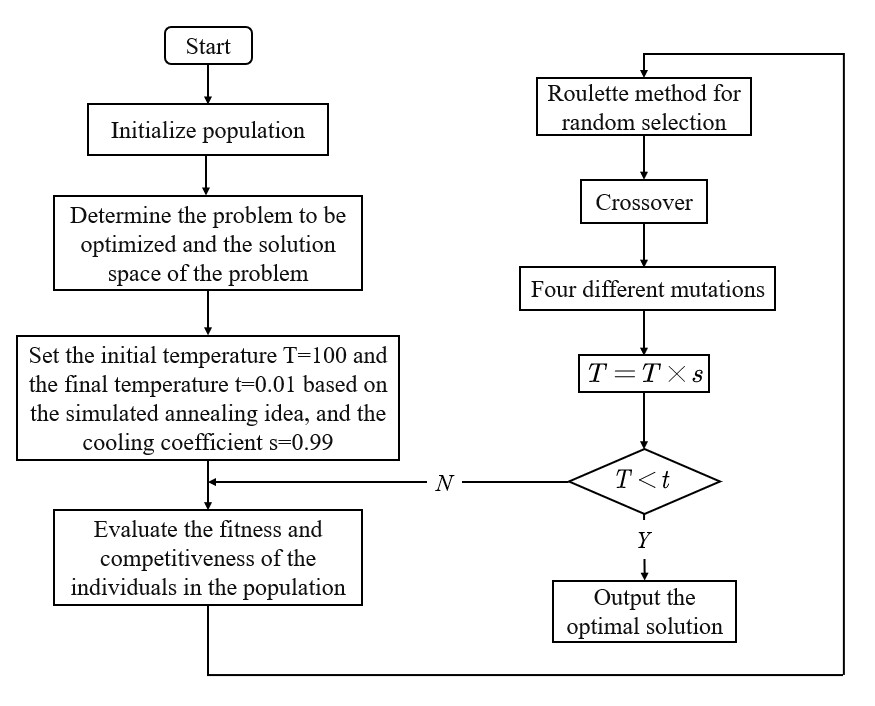} 
\caption{Flowchart of hybrid genetic algorithm combining simulated annealing and greedy methods.} 
\label{GA} 
\end{figure}

In order to solve the difficult problem of training and working with vascular robots, this paper builds a model and then formulates a comprehensive algorithm based on the greedy idea of using simulated annealing to improve the genetic algorithm to solve this problem. Genetic algorithm \cite{srinivas1994genetic} is a computational model that simulates the mechanism of natural selection and inheritance in biological evolution according to the law of biological evolution. This algorithm seeks the optimal solution by simulating the natural evolutionary process. That is, the process of seeking optimal solutions is transformed into the process of chromosome crossover and gene mutation in biological evolution. When solving more complex combinatorial optimization problems, better optimization results can often be obtained faster than traditional optimization algorithms.

Simulated annealing algorithm \cite{van1987simulated} is a generalized stochastic algorithm whose idea is a probabilistic algorithm derived from the process of annealing a solid. A solid at a sufficiently high temperature is slowly cooled and the particles become ordered from disorder. The approximate solution is continuously optimized as the temperature is reduced. The cooling process is controlled in this problem by simulated annealing ideas combined with a genetic algorithm in order to facilitate global optimization. The specific steps of the annealing idea in the genetic algorithm are as follows. The flowchart of the hybrid genetic algorithm combining simulated annealing and greedy methods is shown in Fig~\ref{GA}.
 
	\textbf{Step 1}: Set the initial temperature, T, the temperature change coefficient s, and the equilibrium temperature t.
	
	\textbf{Step 2}: Calculate the objective function values and individual fitness and competitive superiority, and perform optimal solution substitution when a solution that better meets the requirements appears. To converge to the global optimum as much as possible, whenever a more optimal solution appears, the current temperature is warmed up to increase the number of iterations and search breadth \cite{song2020study}. The warming function is $T=T+0.5\ln \left( T-1 \right)$.
                                                                                          
	\textbf{Step 3}: Multiple mating and mutation of the parents after roulette selection to produce offspring, with the range of mutation depending on the current temperature.
	
	\textbf{Step 4}: Add the offspring to the current optimal solution and reduce the current temperature,$T=T\times s$, where $s$ is the cooling coefficient. If the current temperature is not reduced to the termination temperature, repeat \textbf{Step 2} to \textbf{Step 4}. The termination temperature set in this paper is 0.01. When the experimental temperature is reached, the experiment is terminated.
	
	\textbf{Step 5}: Output the global optimal solution at this point after the current temperature is lower than the termination temperature.

\begin{algorithm}[htbp!]
\DontPrintSemicolon
\caption{Improved Genetic Algorithm Based on Greedy and Simulated Annealing}
\label{Al1}
\KwIn{$v$, number of operators and boats purchased, cumulative price impacts from these purchases}
\KwOut{Optimal solution}
sum $\leftarrow$ 0 \tcp*{Total cost of the current solution}
v $\leftarrow$ 0 \tcp*{Total volume of operations for the current solution}
label $\leftarrow$ 0 \tcp*{Label of the current solution}

\SetKwFunction{FGreedy}{Greedy}
\SetKwFunction{FHeat}{Heat}
\SetKwFunction{FCrossover}{Crossover}
\SetKwFunction{FUpdate}{Update}
\SetKwProg{Fn}{Function}{:}{}
\Fn{\FGreedy{$v$, number of operators, number of boats, price impacts}}{
    \For{$i \leftarrow 1$ \KwTo $m$}{
        label $\leftarrow$ k \tcp*{k is the number of robots}
        \If{sum $>$ sumb}{
            \While{sum $>$ sumb}{
                min $\leftarrow$ inf \tcp*{Minimum cost}
                \For{$j \leftarrow 1$ \KwTo $n$}{
                    \tcp{Adjust the number of operators and boats}
                    \tcp{Update the cost and the number of operations}
                    \If{sum $<$ min}{
                        min $\leftarrow$ sum
                        m $\leftarrow$ k
                    }
                }
            }
        }
    }
}

\Fn{\FHeat{T, best, f}}{
    \If{f $<$ best}{
        T $\leftarrow$ T $\times$ 1.1 \tcp*{Warm up}
        best $\leftarrow$ f
    }
}

\Fn{\FCrossover{parent, T}}{
    \tcp{Select two parents and perform crossover and mutation}
    \KwRet{child}
}

\Fn{\FUpdate{T, child, best}}{
    \tcp{Update solution and temperature}
}

\While{T $>$ 0.01}{
    \tcp{Roulette selection, mating, and mutation}
    \tcp{Warm up, update solution}
}

\tcp{Output the optimal solution}

\end{algorithm}

The two algorithms are combined in this problem in order to solve the global optimal solution more accurately and with higher confidence. The flow of the genetic algorithm based on the simulated annealing idea after optimization is shown below:

Greedy algorithms \cite{temlyakov1998best} refer to solving a problem by always making the choice that seems best at the moment. That is, the algorithm obtains a locally optimal solution under the relevant constraints without considering the overall optimality. In this paper, we adopt the greedy idea of minimum cost, from a feasible solution that satisfies the hospital's treatment volume, we continuously reduce the number of robot parts purchased, in order to achieve the purpose of cost reduction, and we hope that we can finally obtain the optimal solution of the problem through the greedy choice of the number of robots purchased each time. In addition, considering that the data is too cumbersome and belongs to the NP problem, the time complexity of linear programming is too high, so it is considered that the initial optimal solution can be obtained through the greedy idea first, and the preliminary optimal solution obtained by this greediness is used as the initial solution in the genetic algorithm based on the simulated annealing idea. Algorithm~\ref{Al1} embodies the ideas and solutions of the greedy algorithm and the genetic algorithm improved by simulated annealing. In this paper, we adopt a greedy algorithm that aims to minimize costs by continuously reducing the number of robot parts purchased from a feasible solution that satisfies the hospital's treatment volume. The greedy algorithm terminates when a feasible solution is obtained. This preliminary optimal solution serves as the initial solution for the genetic algorithm based on the simulated annealing idea, which further optimizes the solution.

By restructuring and optimizing the algorithm, we conduct a series of experiments on the improved algorithm. Ultimately, it can be concluded that our algorithm has superior performance and accuracy for such application scenarios.

\section{Experiments and Discussions}
\subsection{Experiments}
\begin{figure}[!ht] 
\centering 
\caption{Result by using GA with a greedy algorithm, including specific week, number of vessel boats purchased,	number of operators purchased, number of operators maintained, number of vessel boats maintained,	number of operators involved in training (including "skilled" and "novice" workers) and total cost} 
\includegraphics[width=1\textwidth]{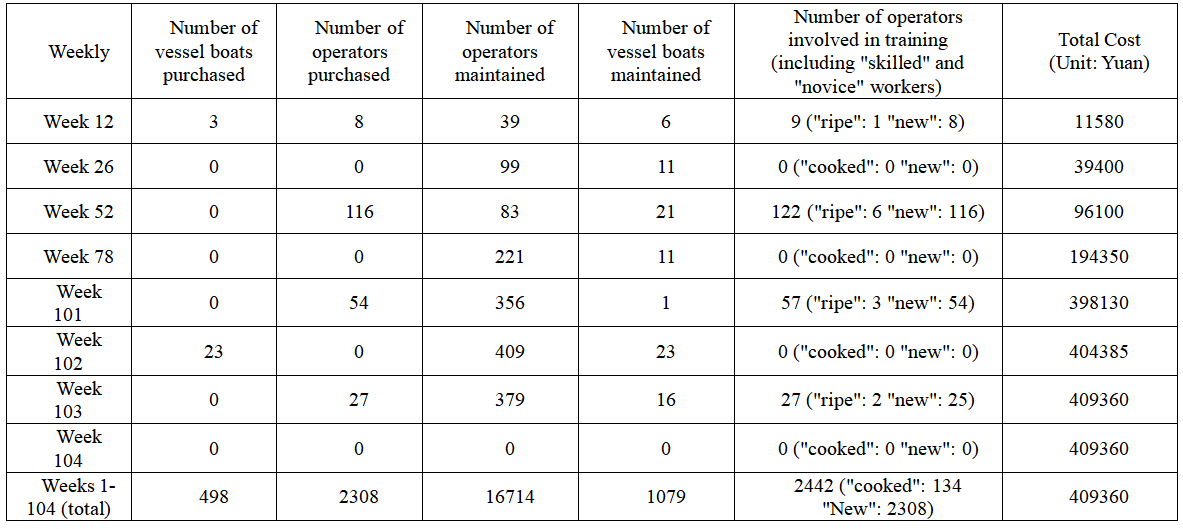} 
\label{table1} 
\end{figure}

\begin{figure}[!ht] 
\centering 
\caption{Result by using GA without the greedy algorithm, including specific week, number of vessel boats purchased,	number of operators purchased, number of operators maintained, number of vessel boats maintained,	number of operators involved in training (including "skilled" and "novice" workers) and total cost} 
\includegraphics[width=1\textwidth]{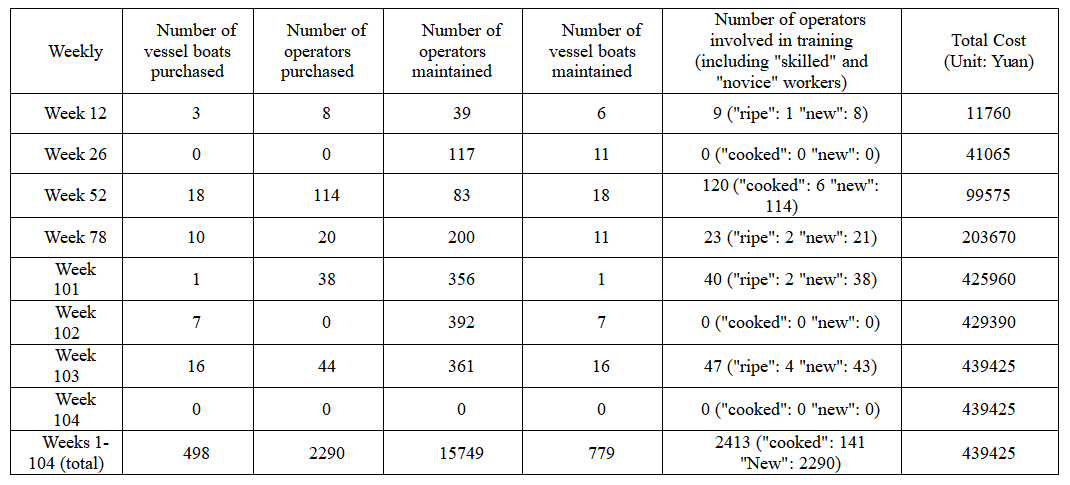} 
\label{table2} 
\end{figure}

The initial optimal solution is first solved by Matlab using the greedy algorithm, and then the initial optimal solution is passed to the genetic algorithm based on the idea of simulated annealing as its initial solution, and then the solution is solved  \cite{yu2023vascular}. Specific purchase data are shown in Fig~\ref{table1} and Fig~\ref{table2}. When only the constants corresponding to the prices of the purchased robot parts are changed in Matlab, and the same genetic algorithm based on the idea of simulated annealing after optimization is used to solve the problem, the solution results are the 498 container boats and 2308 operators purchased in weeks 1-104 will both satisfy the treatment and minimize the cost, with the minimum cost being: 409,360 yuan.

On the genetic algorithm based on the idea of simulated annealing, instead of solving the preliminary optimal solution by the greedy algorithm and then solving it by the genetic algorithm, the algorithm can be used to solve the problem directly as stated earlier. This approach is cost effective and it prepares for the subsequent supply of skilled operators at week 104.

\begin{figure}[!ht] 
\centering 
\includegraphics[width=1\textwidth]{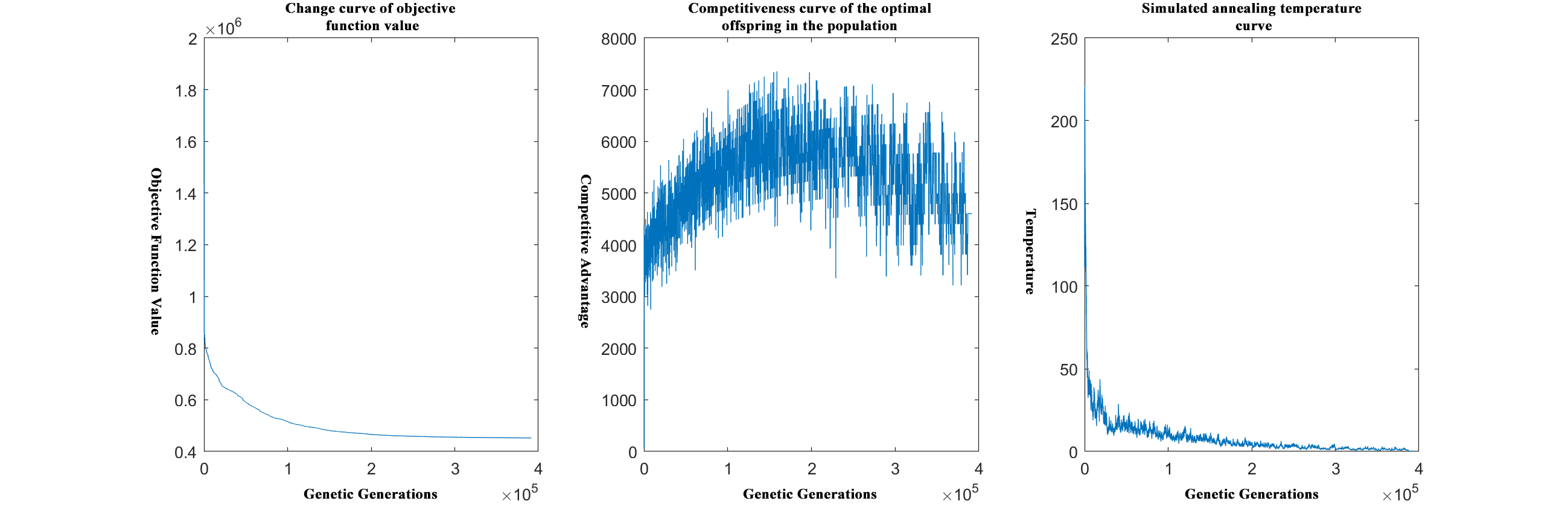} 
\caption{Schematic diagram of simulated annealing and genetic correlation curves} 
\label{result1} 
\end{figure}

Fig~\ref{result1} is a schematic diagram of the relevant curves in the genetic algorithm based on the idea of simulated annealing after optimization, including the way the objective function change curve, the competitiveness curve of the offspring in the genetic algorithm, and the quenching temperature curve in the idea of simulated annealing to help iterate the genetic algorithm.

\begin{figure}[!ht] 
\centering 
\includegraphics[width=1\textwidth]{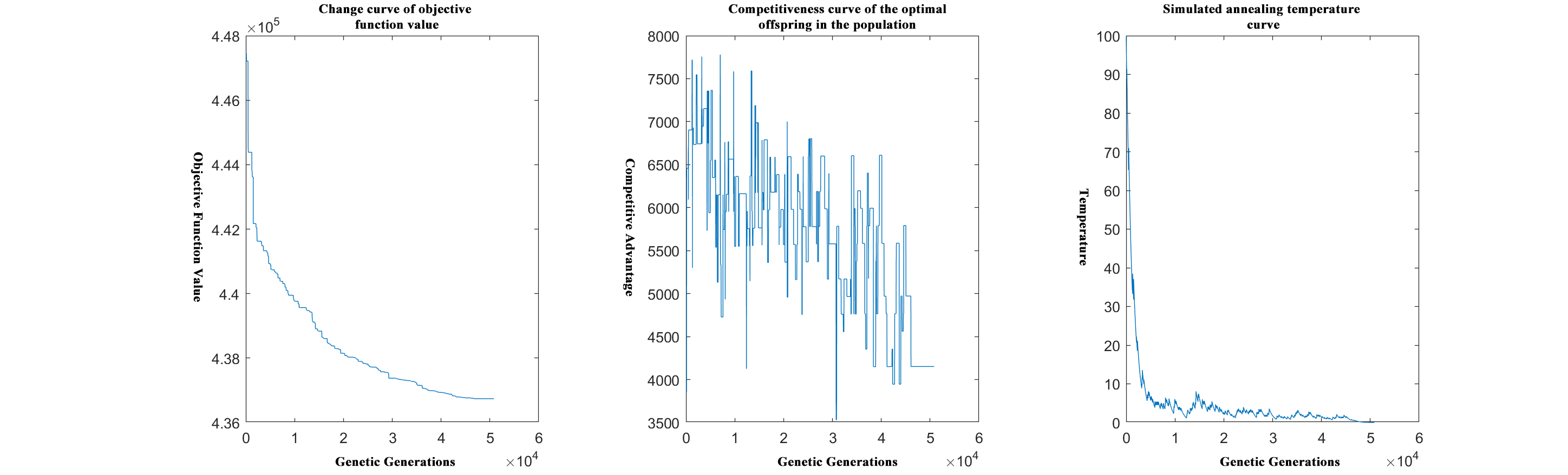} 
\caption{Schematic diagram of the end iterations of the genetic solution after the calculation of the greedy algorithm} 
\label{result2} 
\end{figure}

Fig~\ref{result2} shows the end iterations of the objective function for solving the problem with the greedy algorithm followed by the genetic algorithm.

Obviously, the number of solution iterations of the genetic algorithm is greatly reduced by performing greedy processing first and then performing the genetic algorithm.

\begin{figure}[!ht] 
\centering 
\includegraphics[width=1\textwidth]{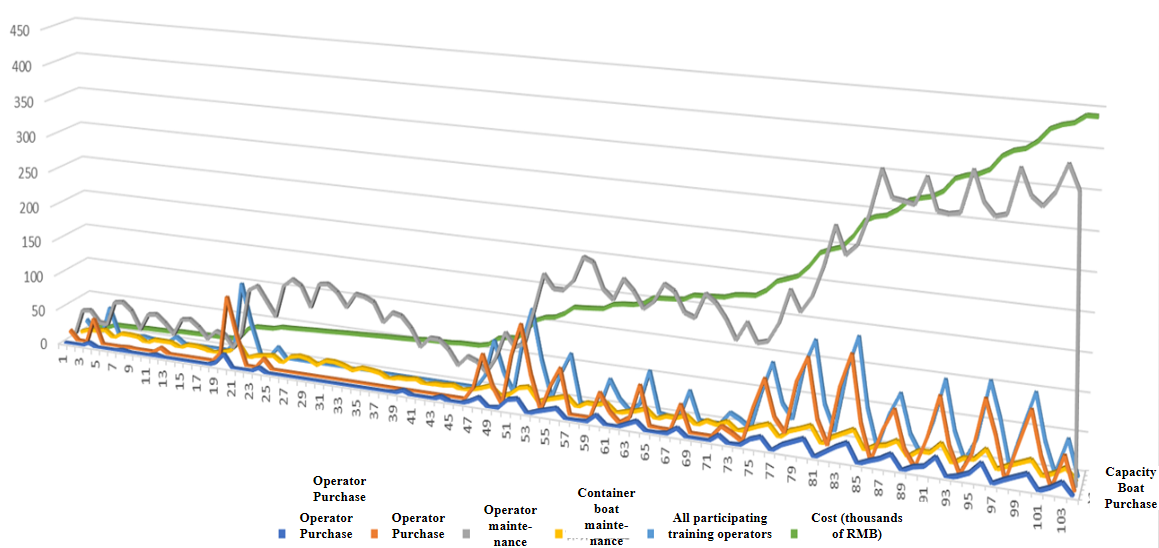} 
\caption{Function image of changes in relevant variables, including capacity boat purchase, operator purchase, operator maintenance, capacity boat maintenance, all operators participating in training, and costs} 
\label{result3} 
\end{figure}

Fig~\ref{result3} shows the images of changes in the variables associated with vessel boats, operators, and weekly costs as a function of.

\begin{figure}[!ht] 
\centering 
\includegraphics[width=1\textwidth]{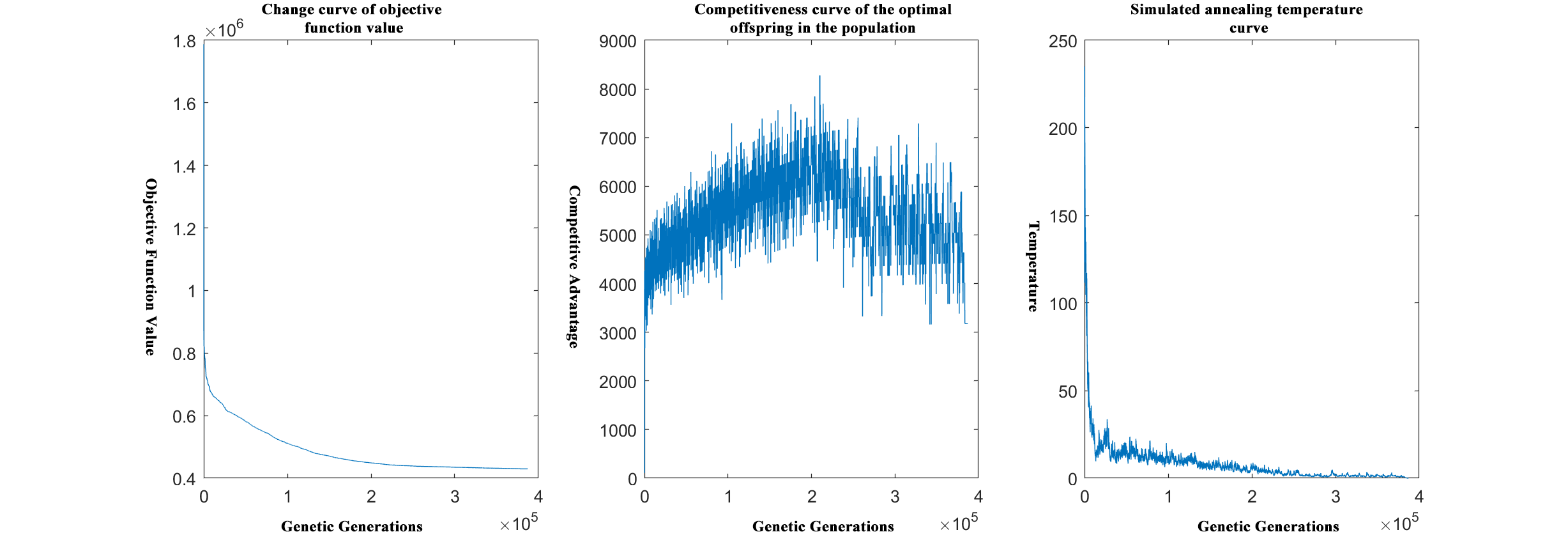} 
\caption{Schematic diagram of simulated annealing and genetic correlation curves} 
\label{result4} 
\end{figure}

When changed the purchase price and the strategy of judging whether to "throw away", that is, the value of the constant and greedy strategy, the main constraints and objective function did not change significantly. Fig~\ref{result4} shows a schematic diagram of the relevant curves in the genetic algorithm based on the idea of simulated annealing after optimization.

\begin{figure}[!ht] 
\centering 
\includegraphics[width=1\textwidth]{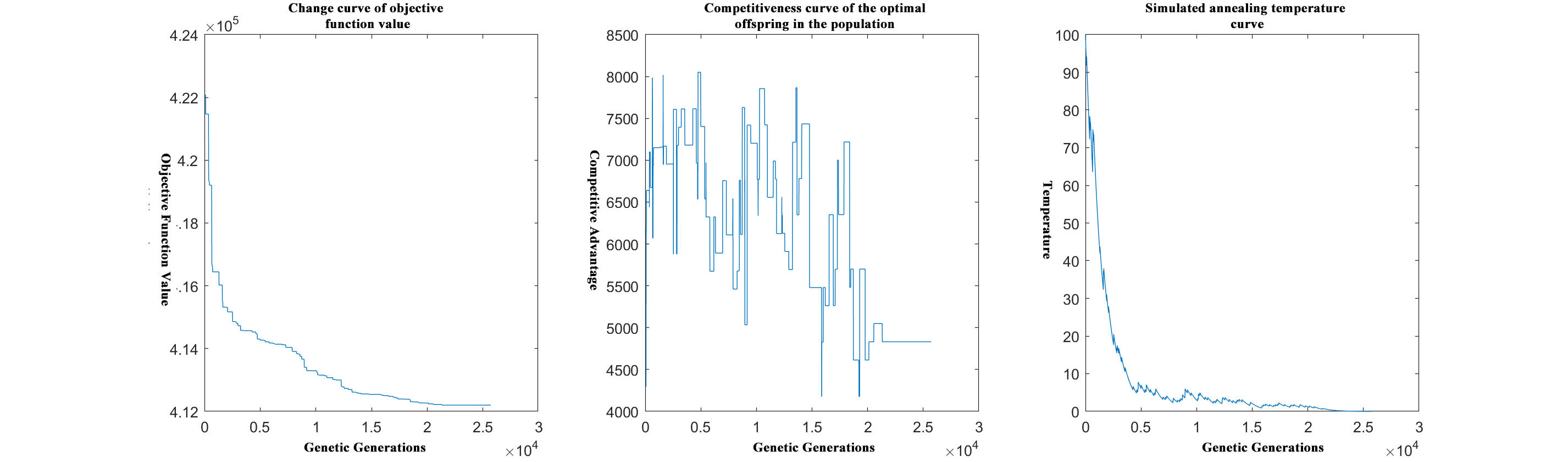} 
\caption{Schematic diagram of the end iteration of the genetic solution after the calculation of the greedy algorithm} 
\label{result5} 
\end{figure}

Fig~\ref{result5} shows that the initial optimal solution is first obtained by the greedy algorithm, and this temporary optimal solution is used as the initial solution in the genetic algorithm model, and then the end iteration of the objective function for solving the problem by improving the genetic algorithm based on the idea of simulated annealing is schematically shown below.

Obviously, greedy processing before the initial optimal solution, and then this solution further into the genetic algorithm model for calculation will reduce the number of iterations will be greatly reduced.

The above graphs reflect the difference in time complexity between calculating the initial optimal solution by the greedy algorithm and solving it directly by the genetic algorithm, which reflects the necessity and importance of the greedy algorithm in solving the model in this paper.

\begin{figure}[!ht] 
\centering 
\caption{ARIMA model parameters} 
\includegraphics[width=1\textwidth]{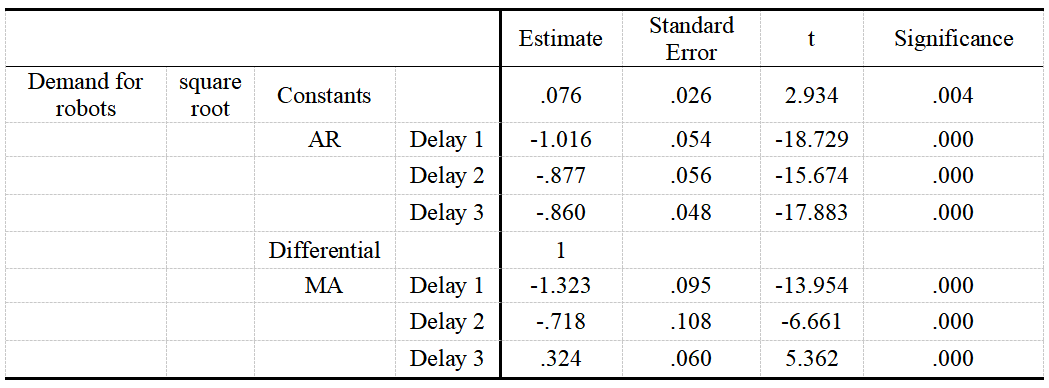} 
\label{table3} 
\end{figure}

Then,a time series analysis of the demand for the use of vascular robots from 1-104 weeks was conducted, and the final model type selected was ARIMA(3,1,4). The ARIMA(3,1,4) model built on the original data was run, and the results are shown in Fig~\ref{table3}. 

From the above table, the coefficients of AR and MA are -1.016, -0.877, -0.860 versus -1.323, -.718, 0.324, respectively.

\begin{figure}[!ht] 
\centering 
\includegraphics[width=1\textwidth]{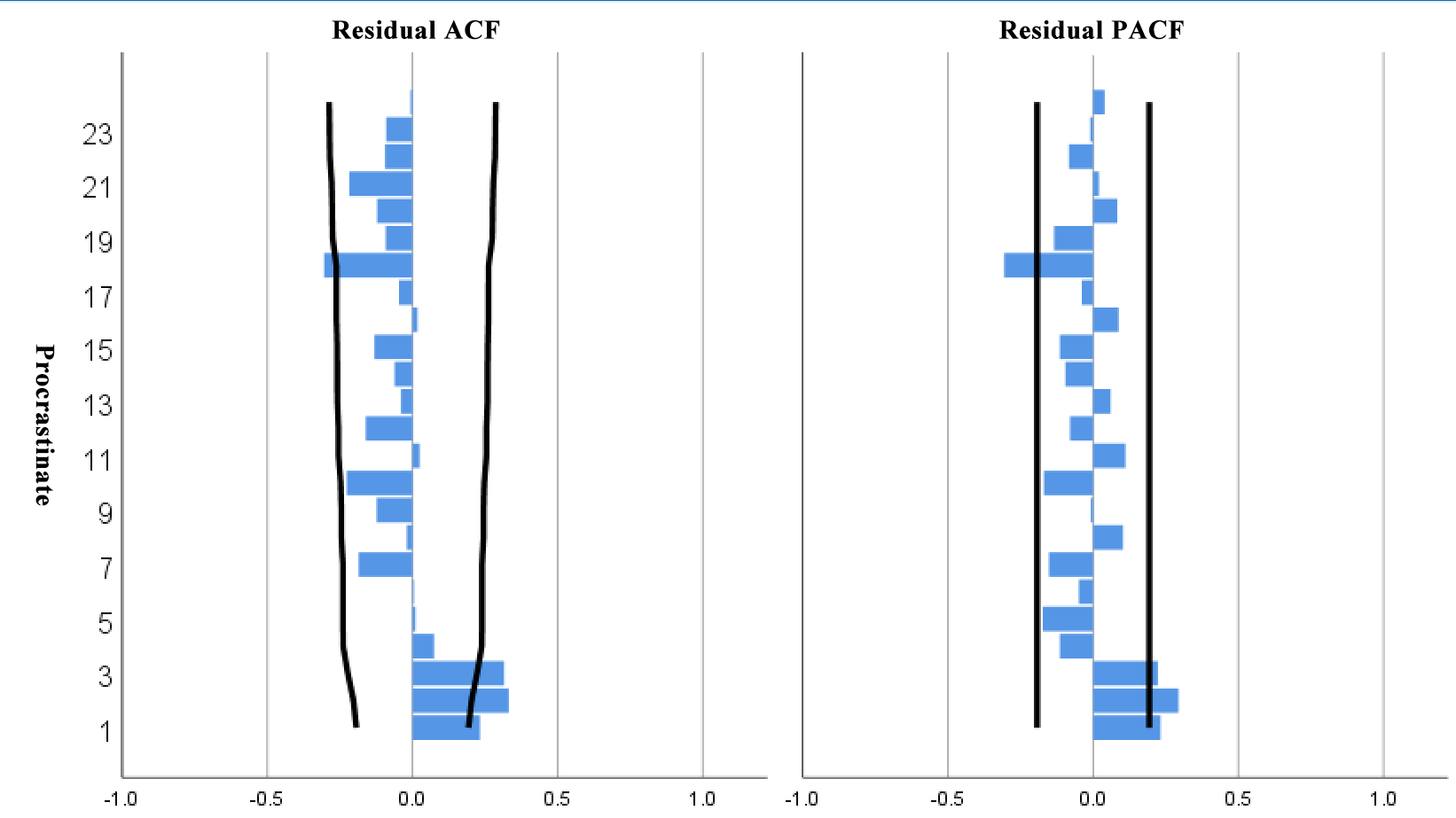} 
\caption{ACF and PACF plots of the residuals. If the ARIMA model has captured all important information about the data, the ACF and PACF plots should resemble white noise.} 
\label{result6} 
\end{figure}

Since the ACF and PACF plots are fluctuating, it is reasonable to choose ARIMA(3,1,4) in this paper. The results of the model are:

\begin{equation}
X_t=\sum\limits_{i=1}^p{\gamma _iy_{t-1}+\epsilon _t+\sum\limits_{i=1}^q{\theta _i\epsilon _{t-1}}}
\label{eq34} 
\end{equation}

Where, $\gamma = [-1.016, -0.877, -0.860] , \theta = [-1.323, -0.718, 0.324]$.

Finally, the goodness-of-fit analysis was performed as Fig~\ref{result6} on ACF and PACF plots of the residuals, and the obtained values related to the model fit are detailed in the Appendix.

The model fit $R_2$ =0.987, which is close to 1, indicating a good fit and a good fitting effect, and the fitted prediction graph is shown as Fig~\ref{result7}.

\begin{figure}[!ht] 
\centering 
\includegraphics[width=1\textwidth]{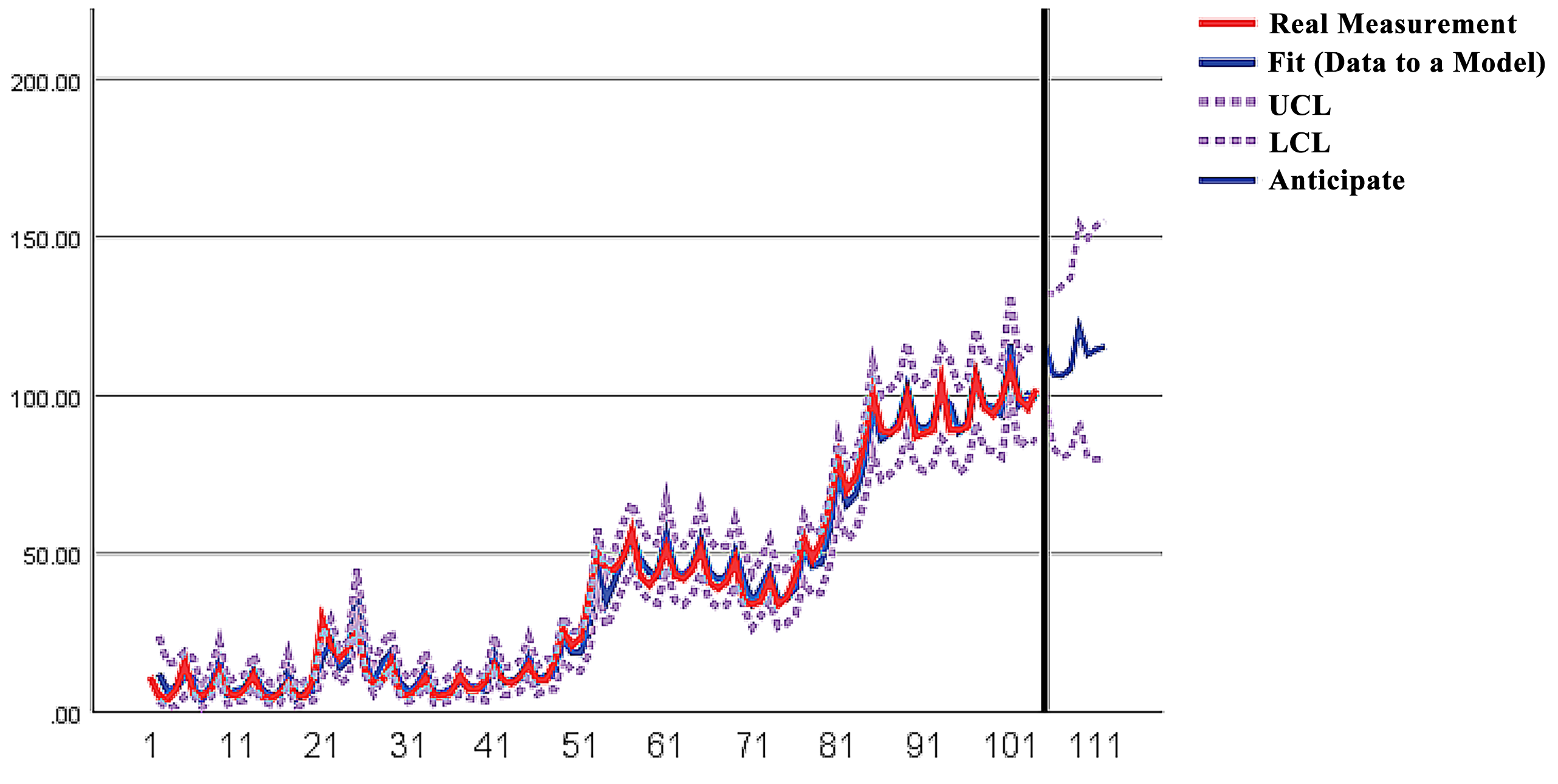} 
\caption{ARIMA model fitted prediction graph} 
\label{result7} 
\end{figure}

\subsection{Discussions}

In the field of healthcare robotics and resource allocation, several algorithms and approaches have been proposed to address similar optimization problems. This section provides a comparative analysis of our proposed algorithm with respect to existing methods, highlighting the advantages and contributions of our approach.

\subsubsection{Traditional Methods} 
To demonstrate our method's efficacy, we conducted comparative analyses with several state-of-the-art approaches, including those by Zhang et al. \cite{zhang2022ordering}, Yu et al. \cite{yu2023vascular}, and Deng et al. \cite{deng2022analysis}. For accuracy, we retrained the models from these studies using the same dataset as ours, evaluating both the number of iterations and total cost. Table~\ref{tab1} indicates that our algorithm significantly surpasses these methods in convergence speed and final cost efficiency.

\begin{table}
\caption{Comparison with SOTA model in number of iterations and final cost} 
\label{tab1}
\centering
\begin{tabular}{ccc}
   \toprule
   Method & Weeks 1-104 (total cost) $\downarrow$ & Number of iterations $\downarrow$   \\
   \midrule
   Zhang et al. \cite{zhang2022ordering} & 436540 & 299631   \\
   Yu et al. \cite{yu2023vascular} & 680080 & 149639   \\
   Deng et al. \cite{deng2022analysis} & 474520 & 196542   \\
   \textbf{Ours} & \textbf{403960} & \textbf{146784}   \\
   \bottomrule
\end{tabular}
\end{table}

In healthcare resource allocation, traditional heuristics, such as greedy algorithms and rule-based policies, have been prevalent. While these methods offer intuitive and computationally efficient solutions, they generally fall short in identifying globally optimal solutions. By contrast, our method employs systematic exploration of the solution space through mathematical optimization techniques, yielding robust, near-optimal results. Therefore, to validate the accuracy and robustness of our method, we compare it with other swarm optimization algorithms, including PSO \cite{marini2015particle}, CGWO \cite{too2021opposition}, GA \cite{srinivas1994genetic}, and AOA \cite{abualigah2021arithmetic}.Since they require simultaneous iterations, this experiment will compare their convergence speed and accuracy. The results of the comparison are shown in Fig~\ref{result7 }.

\begin{figure}[!ht] 
\centering 
\includegraphics[width=1\textwidth]{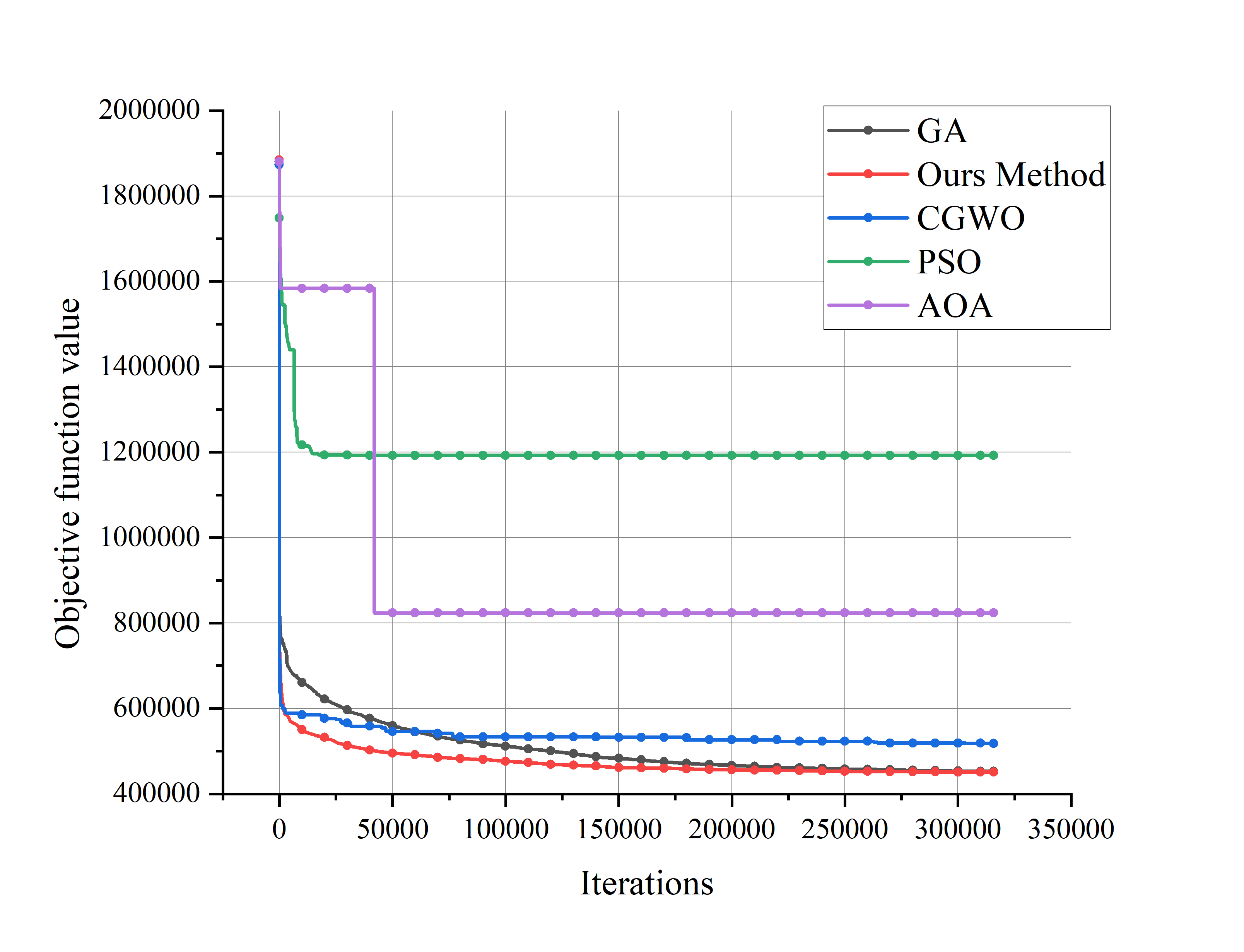} 
\caption{ Investigation on the accuracy and convergence speed of different swarm optimization algorithms} 
\label{result7 } 
\end{figure}

Analysis of the images reveals that the enhanced genetic algorithm, which integrates greedy search and simulated annealing, surpasses competing algorithms in convergence speed and accuracy. This superiority stems from the initial application of the greedy search to rapidly identify a near-optimal solution, which is subsequently refined by the genetic algorithm. This strategy effectively circumvents the issue of entrapment in local optima, a common pitfall in swarm algorithms (e.g., in PSO, AOA curves). Additionally, the greedy search's expedited discovery of an approximate optimal solution, followed by the precision-enhancing simulated annealing, results in the improved genetic algorithm's superior performance in terms of both convergence speed and accuracy.

\subsubsection{Machine Learning Approaches}
Machine learning techniques, including neural networks and reinforcement learning, have been applied to resource allocation problems in healthcare \cite{mizan2022medical}. While these methods can adapt to complex patterns and dynamic environments, they often require substantial amounts of data for training and may not provide interpretable solutions. In contrast, our approach relies on transparent mathematical models, allowing for better understanding and control of the optimization process.

Some recent approaches combine optimization and machine learning components to address resource allocation challenges. While hybrid methods can leverage the strengths of both paradigms, they may introduce additional complexity and computational overhead \cite{tawhid2021machine}. Our algorithm offers a balance by providing efficient optimization while maintaining transparency and ease of implementation. However, we also believe that machine-learning approaches have tremendous potential and room for optimization in the medical field.


\section{Conclusions}

This research have addressed the pressing challenges of optimizing robotic operator and vessel boat acquisition strategies within the dynamic healthcare environment. Our research objectives, including the development of a robust buying strategy model, adaptability to macrophage attacks, consideration of skilled operator variations, and the creation of a comprehensive framework, have all been successfully achieved. This study significantly contributes to healthcare robotics by bridging existing gaps in the literature and offering practical solutions that enhance cost-effectiveness, treatment efficiency, and resource allocation. The potential impact of our research on medical treatment, particularly in vascular diseases and virus removal, is substantial. While this work represents a significant advancement, future research can explore real-world implementation and further incorporate advanced technologies for even greater adaptability and prediction accuracy. Overall, our study marks a pivotal step in the evolution of healthcare robotics, with far-reaching implications for patient care and well-being.

\bibliographystyle{unsrt}  
\bibliography{references}

\end{document}